\documentclass{IOS-Book-Article}

\usepackage{mathptmx}
\usepackage{soul}\setuldepth{article}

\usepackage{tikz}
\usepackage{graphicx}
\usetikzlibrary{shapes.geometric, arrows, calc, fit, positioning, chains, arrows.meta}

\def\hb{\hbox to 11.5 cm{}}

\begin{document}

\pagestyle{headings}
\def\thepage{}

\begin{frontmatter}              

\title{Monitoring AI systems: A Problem Analysis, Framework and Outlook}

\markboth{}{March 2022\hb}

\author[A]{\fnms{Annet} \snm{Onnes}%
\thanks{Corresponding Author: Annet Onnes; E-mail:
a.t.onnes@uu.nl.}},

\runningauthor{A.T. Onnes et al.}
\address[A]{Information and Computing Sciences, Utrecht University, The Netherlands}

\begin{abstract}
Knowledge-based systems have been used to monitor machines and processes in the real world. In this paper we propose the use of knowledge-based systems to monitor other AI systems in operation. We motivate and provide a problem analysis of this novel setting and subsequently propose a framework that allows for structuring future research related to this setting. Several directions for further research are also discussed.
\end{abstract}

\end{frontmatter}
\markboth{March 2022\hb}{March 2022\hb}

\section{Introduction}
When working in a team, expectations exist about how work should be done or how members of the team ought to collaborate. Members of a team have to rely on each other, knowing that expectations are fulfilled. When we consider hybrid teams, where humans and AI systems work together, this includes the artificial team members \cite{Akata2020}. 

An AI system working in a team works in a specific environment. Even when we assume that everything is done to obtain the best possible performance during the process of designing and implementing the system, putting it into use in a specific environment can change what good performance looks like. For example, an AI system that makes treatment recommendations based on patient data can be trained using data from several different hospitals, in order to reach the best possible performance in the general task. However, when the system is used in a particular hospital, it might have to adjust its recommendations based on e.g. the protocols in place in that hospital. The idea is that when a system is deployed in a team or other specific context, measuring performance like in the general training phase might not suffice as a measure of whether a system performs satisfactorily in that specific context. 

In order to measure performance in a specific context, the AI system has to be monitored while it is running. There are many existing perspectives on what monitoring is when regarding using knowledge-based systems for monitoring real-world processes or simulations thereof (see e.g. \cite{DellAnna2018, Chandola2009, Avila2014}). In this paper we present a novel framework in which knowledge would be used to monitor other AI systems, which presents its own challenges and requires its own perspective. We aim to study how to monitor AI systems in such a way that the expectations for the performance are formulated into an interpretable, knowledge-based system for monitoring.

We begin with a conceptual analysis of various concepts such as modelling and monitoring in Section \ref{sec:analysis}. Following this, two main research questions can be formulated. Section \ref{sec:framework} provides a specification of the monitoring setting, which helps to give structure and clarity to research challenges that arise from the posed questions. We present a framework using a design pattern language that covers an AI system and its monitoring system. Section \ref{sec:discussion} discusses further nuances of the research challenges presented in Section \ref{sec:analysis} as supported by the framework in Section \ref{sec:framework}.

\section{Problem Analysis}
\label{sec:analysis}
In this section we introduce the problem setting of monitoring AI systems. We begin the problem analysis by introducing the concepts involved and weaving them together in an overview. We then further unravel two aspects of it in order to add more nuanced considerations and ask specific questions.

\subsection{Problem overview}
\label{sec:overview}
The problem setting consists of an AI system, which has to be monitored because there are expectations about its functioning in a specific context. The expectations provide prescriptions for a monitoring system, which uses some methodology to flag output that does not adhere to the expectations. An impression of this is given in Figure \ref{fig:impression}.

\input{impression_setting}

Using input an \emph{AI system} makes predictions or decisions, which are the systems output. It uses a \emph{model}, which is defined as a simplified representation of some real (data-generating) process. The process on which a model is based is referred to as the \emph{target system} \cite{Frigg2020}. This process can be, for example, the development of the weather about which a model can make predictions. It can also be a decision making process, for example whether a bank will offer a loan. The distribution of all data generated by and collected from a target system is referred to as its \emph{true distribution}. 

\emph{Models} can take many forms, ranging from a machine learning model to a toy railway, they all represent a target system in some way. Although it could be desirable to monitor other types of models, the focus here is on statistical models. A statistical model is trained using parts of the data from the target system that are observable. 





%


Next we have the \emph{monitoring system}, which is, informally, meant to keep an eye on the AI system to see whether the output is looking good in the specific context of operation. We more precisely define it as the system that identifies output of the AI system that does not fulfil the expectations. When this is the case, the output is \emph{flagged}. The monitoring system is not defined as deciding what output is \emph{wrong}; The expectations the output should adhere to are determined externally, meaning monitoring itself is a technical operation devoid of assigning value. A monitoring system can be looking out for whatever it is assigned to look out for. The monitoring system requires external instructions about what the expectations are, we call these \emph{prescriptions}. The prescription is specified by human users so the system is able to monitor according to their expectations in a specific context. This is information that is not included in the AI system because it can be context specific. Prescriptions can for example be demands to adhere to norms or protocols in a collaborative setting. Output that is accurate according to the AI system and that also adheres to the prescription is referred to as \emph{intended behaviour}. Ideally a AI system would have a model of which the target system is the intended behaviour, the overlap of optimal performance and fulfilling expectations.

Both the AI system model and the prescriptions supervene on the target system. Since the AI system is trained to fit the data generated by the target system and human interpretation of the process shapes user expectations, which inform prescriptions.




The following section further examines monitoring, the task of checking whether the AI system output adheres to the prescription, and the research challenges in that task. Section \ref{sec:prescriptions} will further elaborate on the difference between monitoring the adherence to a prescription and determining what the intended behaviour ought to be.

\subsection{Monitoring} 
\label{sec:monitoring} 
To determine how to identify output that does not adhere to the prescription, we first consider research on \emph{anomaly detection}. Anomaly detection techniques are used to identify cases, known as anomalies, that are abnormal cases compared to the data that represents normal behaviour (See Chandola et al. \cite{Chandola2009} for a survey.) Anomaly detection can help, for example, to look for fraud in credit card use by considering data like type of exchanges, amounts, location, time and frequency of use. In general in anomaly detection, the way of representing normal behaviour is through a model based on data (see e.g. \cite{Chandola2009,Mascaro2014,Kirk2014}). This means this model is \emph{descriptive}, it describes the process that produced the normal behaviour data. 

The issue at hand here is slightly different. Anomaly detection techniques that are descriptive aim to detect anomalies using a model based on data of normal behaviour in a target system. In our case we need to compare the in- and output of the AI system, i.e. observed model behaviour, to intended behaviour. We have to reconsider how to represent this intended behaviour because modelling the intended behaviour using data, like modelling normal behaviour, is ineffective. Doing this would result in another model of a target system, similar to the AI system's model. This means a different representation is required, for instance a knowledge-based model constructed using the prescriptions.


We thus have two research questions, \emph{how to do the comparison between the intended behaviour and the observed AI system behaviour?} and \emph{how to represent the intended behaviour using prescriptions?} Anomaly detection research can help with the first, as it presents ways of comparing specific cases with normal behaviour. These methods often rely on measures for probabilities and distributions to detect outliers as being anomalous.


The probability of a case, given a distribution over cases, can be used as a measure \cite{Johansson2007}. This considers `Is this instance unlikely?' However, being \emph{unlikely} does not necessarily indicate being an outlier. Especially considering that the more possibilities there are, the smaller the probabilities will be. To address this particular issue, a conflict measure was introduced, which considers the consistency within a case consisting of multiple values (an instance) \cite{Nielsen2007}. 

Aside from measuring the likelihood of instances, measures have been developed that compare two distributions. A Bayes factor is a measure that compares the likelihood of one instance in two distributions \cite{Kass1995}. Examples of comparing entire distributions using likelihood are relative entropy, also known as the Kullback-Leibler divergence \cite{Kullback1951} or the CD Distance measure \cite{Chan2005}, which addresses shortcomings of the KL-divergence. They are of limited use in this context since they both require to run the AI system and any prescription model we might have, for each possible outcome in order to calculate them. Unless they are calculated analytically which requires full access to the AI systems, which we do not have.



Such an analytical approach would lead to an a priori way of monitoring, which tells in advance whether a model would fulfil expectations \cite{DellAnna2021}. However, this setting monitors during runtime which also avoids checking every possible input. Rather than identifying all unintended behaviour the AI system \emph{could} show a priori, it is more effective to have to infrequently flag behaviour during run time. A run-time approach would also be beneficial to further investigation into monitoring continual learning AI systems, as continual learning means continual changes to the model.

\subsection{Prescriptions and intended behaviour}
\label{sec:prescriptions}
It is one thing to keep an eye on the AI system to see whether the output is adhering to expectations, but another to make decisions about what the expectations are and thus whether the output `good' or `bad'. 

In order to make those decisions it has to be known what is intended, like it has to be known for anomaly detection what normal behaviour is. However, when e.g. monitoring an AI system is used for loan decisions to make sure its decisions are and continue to be fair, it is unclear what fair would entail \cite{Tsamados2022}. What is considered fair is not necessarily represented in the training data and therefore neither in the model that the system uses. The monitoring system requires additional information about the fairness of decisions, aside from the information in the AI system. This section considers norms as a sketch of such information.

In social situations, like team work, \emph{norms} are expectations of behaviour. Therefore norms could help prescribe what a monitoring system needs to look out for. Norms could provide an answer to the question what \emph{should} be done as they give an indication of intended behaviour. In Brennan et al. \cite{Brennan2013} a norm is characterised as a particular rule or normative principle that is accepted in a particular group or community. Similarly to Brennan et al. we are not interested in norms as statistical norms -- which would be normal behaviour -- or an objective normative principle -- which is without social context. Our preferred characterisation of norms consists of two parts, 1) a normative principle dictating \emph{what} should be done or not and 2) a \emph{socio-empirical} dictation of \emph{who} that applies to \emph{when} \cite{Brennan2013}. Dastani et al. refer to a norm as ``an expected behaviour in a social setting'' \cite{Dastani2018}. This matches the description from \cite{Brennan2013}, because expected behaviour is the normative principle and the social setting is a socio-empirical dictation. \emph{Norm monitoring} is then described as ``determining whether an individual is adhering to this expected behaviour''. This fits the characterisation of monitoring in Section \ref{sec:overview}.

Suggesting norms as the prescription used by monitoring systems does not answer the question what the AI system should be doing, nor does it provide a representation of prescriptions for the monitoring system. Norms are suggested as a potential partial representation of intended behaviour aside from the AI system model because they are indicators of the expectations of humans. Norms as specified rules, for example in the case of protocols, are a more concrete step towards a representation. The research challenge is to further determine how norms can inform the monitoring system about the intended behaviour in such that more information is accessible for the comparison between the observed system and intended behaviour

\section{Framework Description}
\label{sec:framework}
The aim of this section is to provide clarity about the monitoring setting using a structured formalisation, following the problem analysis from the previous section. While in subsection \ref{sec:overview} the AI system, monitoring system and prescription were introduced and further nuance into these elements was provided, the overall setting of monitoring can be made clearer. This section uses modular design patterns to formalise the framework sketched in Figure \ref{fig:impression}.


\subsection{Framework Design Pattern}
In order to further formalise the framework, we will refer to the modular design patterns and taxonomical vocabulary introduced by Bekkum et al. \cite{Bekkum2021}, which were proposed in order to describe in a unifying way the architecture of a whole range of systems that combine statistical and symbolical methods. This means it provides the vocabulary to describe this monitoring setting in an abstract framework.

The basis of the vocabulary consists of four elements: \begin{tikzpicture}

    \tikzstyle{every node}=[]
	
    \tikzset{actor/.append style={draw, black, shape=rectangle, double}}
    \tikzset{process/.append style={draw, black, shape=rectangle, rounded corners=0.25cm}}
    \tikzset{instance/.append style={draw, black, shape=rectangle}}
    \tikzset{model/.append style={draw, black, shape=rectangle, fill=gray!30}}

    \tikzstyle{every path}=[draw, thick]

    \node[instance] (E) at (4,2) {\texttt{instance}};

\end{tikzpicture}, \begin{tikzpicture}

    \tikzstyle{every node}=[]
	
    \tikzset{actor/.append style={draw, black, shape=rectangle, double}}
    \tikzset{process/.append style={draw, black, shape=rectangle, rounded corners=0.25cm}}
    \tikzset{instance/.append style={draw, black, shape=rectangle}}
    \tikzset{model/.append style={draw, black, shape=rectangle, fill=gray!30}}

    \tikzstyle{every path}=[draw, thick]

    \node[model] (E) at (4,2) {\texttt{model}};

\end{tikzpicture}, \begin{tikzpicture}

    \tikzstyle{every node}=[]
	
    \tikzset{actor/.append style={draw, black, shape=rectangle, double}}
    \tikzset{process/.append style={draw, black, shape=rectangle, rounded corners=0.25cm}}
    \tikzset{instance/.append style={draw, black, shape=rectangle}}
    \tikzset{model/.append style={draw, black, shape=rectangle, fill=gray!30}}

    \tikzstyle{every path}=[draw, thick]

    \node[process] (E) at (4,2) {\texttt{process}};

\end{tikzpicture} and \begin{tikzpicture}

    \tikzstyle{every node}=[]
	
    \tikzset{actor/.append style={draw, black, shape=rectangle, double}}
    \tikzset{process/.append style={draw, black, shape=rectangle, rounded corners=0.25cm}}
    \tikzset{instance/.append style={draw, black, shape=rectangle}}
    \tikzset{model/.append style={draw, black, shape=rectangle, fill=gray!30}}

    \tikzstyle{every path}=[draw, thick]

    \node[actor] (H) at (0,2) {\texttt{actor}};

\end{tikzpicture}. Elements can be further specified using a colon \texttt{:}, for example a model can be statistical or semantic (e.g. \texttt{model:stat} or \texttt{model:sem}). These elements can be combined to form design patterns. The simplest are elementary design patterns which describe one process with in- and output. Boxes \texttt{1} and \texttt{2} in Figure \ref{fig:framework} represent design patterns of the general processes for e.g a knowledge base being engineered or a machine learning model being trained, respectively.




Using the design patterns, we can formally depict the systems and processes of the monitoring setting, as sketched in Figure \ref{fig:impression}. The framework shown in Figure \ref{fig:framework} represents all elements from Section \ref{sec:overview} that are designed systems with in- and outputs. It therefore does not include the target system as in Figure \ref{fig:impression}.

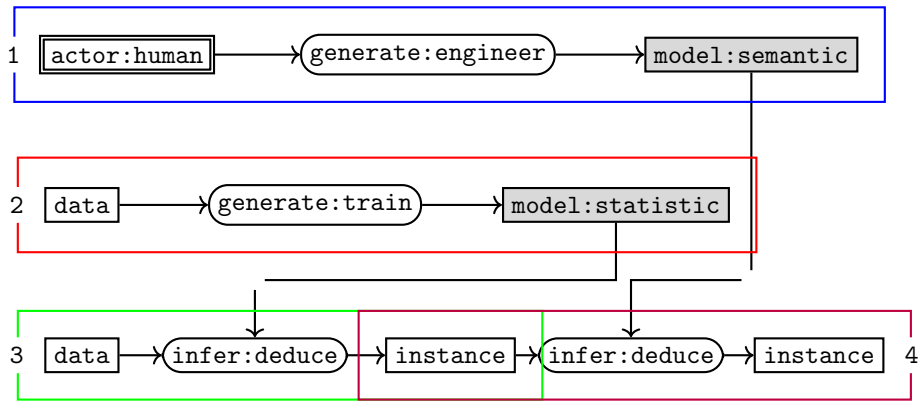
\begin{figure}[h]
\begin{tikzpicture}

    \tikzstyle{every node}=[]
	
    \tikzset{actor/.append style={draw, black, shape=rectangle, double}}
    \tikzset{process/.append style={draw, black, shape=rectangle, rounded corners=0.25cm}}
    \tikzset{instance/.append style={draw, black, shape=rectangle}}
    \tikzset{model/.append style={draw, black, shape=rectangle, fill=gray!30}}

    \tikzstyle{every path}=[draw, thick]

    \node[actor] (H) at (0,2) {\texttt{actor:human}};
    \node[process] (E) at (4,2) {\texttt{generate:engineer}};
    \node[model] (G) at (8.3,2) {\texttt{model:semantic}};

    \path [->] (H) -- (E);
    \path [->] (E) -- (G);

    \node[instance] (D) at (-0.6,0) {\texttt{data}};
    \node[process] (T) at (2.5,0) {\texttt{generate:train}};
    \node[model] (M) at (6.5,0) {\texttt{model:statistic}};

    \path [->] (D) -- (T);
    \path [->] (T) -- (M);

    \node[instance] (I) at (-0.6,-2) {\texttt{data}};
    \node[process] (R) at (1.7,-2) {\texttt{infer:deduce}};
    \node[instance] (O) at (4.3,-2) {\texttt{instance}};
    \node[process] (C) at (6.7,-2) {\texttt{infer:deduce}};
    \node[instance] (F) at (9.2,-2) {\texttt{instance}};

    \path [->] (I) -- (R);
    \path [->] (R) -- (O);
    \path [->] (O) -- (C);
    \path [->] (C) -- (F);

    \node[] (H1) at (1.7,-1) {};
    \node[] (H2) at (8.3,-1) {};

    \path [-] (M) |- (H1);
    \path [->] (H1) -- (R);
    \path [-] (G) -- (H2);
    \path [->] (H2) -| (C);

    \node[draw=blue, thick, inner xsep=1em, inner ysep=1em, fit=(H) (E) (G)] (box) {};
    \node[draw=white, fill=white] at (box.west) {\texttt{1}};

    \node[draw=red, thick, inner xsep=1em, inner ysep=1em, fit=(D) (T) (M)] (box) {};
    \node[draw=white, fill=white] at (box.west) {\texttt{2}};

    \node[draw=green, thick, inner xsep=1em, inner ysep=1em, fit=(I) (R) (O)] (box) {};
    \node[draw=white, fill=white] at (box.west) {\texttt{3}};

    \node[draw=purple, thick, inner xsep=1em, inner ysep=1em, fit=(O) (C) (F)] (box) {};
    \node[draw=white, fill=white] at (box.east) {\texttt{4}};

\end{tikzpicture}
\caption{The full monitoring setting with AI system and monitoring system. Marked are four individual sections which are elementary patterns.}
\label{fig:framework}
\end{figure}

In Figure \ref{fig:framework} four processes are marked and numbered. The elementary pattern \texttt{1} describes the process through which prescriptions are modelled from human knowledge. The \emph{Humans} and \emph{Expectations} elements in Figure \ref{fig:impression} also refer to this process. The resulting semantic model informs the monitoring system (represented by pattern \texttt{4}). Box \texttt{2} is the elementary pattern for training a statistical model from data. In Figure \ref{fig:impression} it is the step from \emph{Target system} to \emph{AI system}, although here only the data is represented. \texttt{model:statistic} is the model used by the AI system represented by pattern \texttt{3}, that makes predictions or decisions. Box \texttt{4} is the monitoring process, which is a deductive process, using the AI system's predictions (and input) as input.


The framework gives an explicit formalisation of a monitoring setting. With this, any further research can be structured more easily, whether considering specific elements, processes or the relation between them. For instance, consider the frequency at which each of these processes is run. The patterns in Figure \ref{fig:framework} are presented on three different levels, where each level can have its own frequency. The frequency is lowest when a model is created a priori after which it is used statically and the frequency is highest at run-time, when a system is constantly running to produce new output. The prescription model is created a priori by humans, in process \texttt{1}. At the level of \texttt{3} and \texttt{4} the frequency is run-time. This leaves the question of the frequency at which process \texttt{2} is run. In Section \ref{sec:discussion} we further cover how this framework can be used to study matters including these.


\section{Discussion and Future Work}
\label{sec:discussion}
The groundworks in Sections \ref{sec:analysis} and \ref{sec:framework} are only the stepping stones from which to continue research in this monitoring setting. In this section we further discuss: the role of uncertainty in modelling and about target systems; the challenge of using distance measures for monitoring and the impact of AI systems starting to learn. 

\subsection{Uncertainty about the target system}
In Section \ref{sec:monitoring} it was discussed that, when the target system and AI system are two distributions they could be compared using (distance) measures. However, this is not a complete method for a comparison. Most prominently because the target system is not accessible. We can not \emph{know} what it is, but the model in the AI system and the prescriptions supervene on it. Knowing these puts constraints constraints on what the intended behaviour -- that is the target system -- could look like. Thus as we try to find a way to compare the intended behaviour and the observed AI system behaviour, we are led to the question of representation. We need a way to represent what we \emph{do know} about the target system and preferably any such representation would make comparison to the AI system behaviour possible.

To illustrate, consider firstly a hypothetical complex system that has completely and correctly modelled the target system and secondly a simplified system where the model is a slimmed down version of the first model. The complex system is equivalent to the target system distribution and the simplified system to the AI system. These systems will not behave identically, the simplified system will on occasion give divergent output. In reality there cannot be such a complex system, but the AI system is still a simplified version. Only, we do not know of what complex system exactly. There is a set of complex systems that the AI system can be the slimmed down version of. Returning to the aim of the monitoring system, it is to identify the outputs from the simplified system that do not match those the intended behaviour. To monitor, given a set of complex systems, one of which is the actual target system, we would need to compare the AI system output to a set of distributions.

We need a way to define and represent, for monitoring purposes a set of complex systems, all possible target systems. This entails figuring out how to use what we \emph{do know} about the target system in order to define the set. Although the AI system and prescriptions put constraints on what the intended behaviour is, it becomes a set of systems because these do not reduce all uncertainty. They constrain what the actual target system could be, but cannot exactly specify it. Every target system exists in the real world in such a way that it cannot be modelled completely, things are left out of a model's simplified representation.  This also holds for the AI system and the prescription, they only partly model the target system. Any uncertainty about the target system that falls outside either of these models remains. This means the actual target system is one of set of possibilities, each simplifying to become the AI system model as well as upholding the prescriptions. This set of is thus a representation of the target system.


\subsection{Measures for Monitoring}
Figure \ref{fig:framework} shows that the prescriptions and AI system model are not integrated, instead the output from the AI system is monitored using the prescriptions in another process. If monitoring can be done by comparing the output and intended behaviour, we must consider what exactly the measures from Section \ref{sec:monitoring} measure and thus monitor for. When measuring abstract concepts, such as the difference between observed AI system behaviour and intended behaviour, different measures can be defined, because there are multiple ways in which things like distributions can be differentiated. Jitkrittum et al. \cite{jitkrittum2018} demonstrate this as they compare a distance and discrepancy measure by calculating these measures for distributions for which the ways in which they differ are known and analyse how this shows in the measurement results. Distributions might differ in density but not in shape and vice versa. To this end we can review various measures and what purpose they might serve best, as it should be possible to monitor without a difference measure that encapsulates all aspects.

\subsection{The AI system as a learning system}
We defined an AI system as modelling a process and being able to output predictions or decisions in Section \ref{sec:analysis}. These outputs depend on input. The system would be a learning system if between different occasions of the system getting the same input, the output could change. This happens when the model is updated, re-trained or changed through some other process. Learning during runtime can be advantageous, but it challenges guarantees about performance. Monitoring is therefore especially helpful in situations where the model can change. To consider this further we need to be aware of how a system can change or what else can change accordingly.

If the AI system can continue to learn during run-time, this means using information from runtime in training. This can be expressed in Figure \ref{fig:framework} by adding an arrow from the output of process \texttt{3} to the learning process in pattern \texttt{2}. Before adding this arrow to the framework, all vertical arrows were pointing down, which meant that the frequency of running processes on the lower levels could be higher (runtime or closer to it) than levels above it. Adding the arrow point upstream means that running process \texttt{2} has to be frequent enough that the learning will take place. Connecting the levels bidirectionally forces the two processes to run at similar frequencies and thus the model from \texttt{2} can no longer be \emph{static} input to \texttt{3}. Even if \texttt{2} does not update for every output of \texttt{3}, in order for learning to happen the AI system's model has to be updated with the new training data sufficiently frequent.







\section{Conclusion}

This paper presented a problem analysis of using knowledge-based models to monitor whether an AI system is behaving according to expectations in a particular context. This analysis helped to identify research questions: \emph{how to do the comparison between the intended behaviour and the observed AI system behaviour?} and \emph{how to represent the intended behaviour using prescriptions?} These questions can be tackled in various ways, leading to different research directions in this setting. Using the abstract framework of the setting presented here these challenges can be formulated in a structured manner. Like with most research there are a vast amount of possible research directions and factors to take into consideration. Not all are mentioned here, such as the nuanced role of types of uncertainty in modelling or the possible representations that would help to integrate norms in the framework. The next step in this endeavour will also include finding technical representations and notations to begin defining the difference between intended behaviour and observed model behaviour.

\section*{Acknowledgements}
This research was funded by the Hybrid Intelligence Center, a 10-year programme funded by the Dutch Ministry of Education, Culture and Science through the Netherlands Organisation for Scientific Research, \url{https://hybrid-intelligence-centre.nl}.

\bibliographystyle{vancouver}
\bibliography{references}

\end{document}